\documentclass[preprint, review, 12pt]{elsarticle}

\usepackage{amssymb}
\usepackage{amsmath}
\usepackage{caption}
\usepackage{booktabs}
\usepackage[margin=1in]{geometry}
\usepackage{graphicx}
\usepackage{multirow}
\usepackage{rotating}
\usepackage{tabularx}
\usepackage{amsfonts}
\usepackage{mathtools}
\usepackage{xcolor}
\usepackage{fancyhdr}		
\usepackage{enumitem}      
\definecolor{red}{rgb}{0.85,0,0}
\definecolor{blue}{rgb}{0,0,0.85}

\DeclareMathOperator*{\argmin}{arg\,min}

\usepackage{graphicx}
\usepackage{subcaption}
\usepackage{lipsum}
\usepackage{makecell}
\usepackage{hyperref}
\hypersetup{ colorlinks,
    citecolor=blue,
    filecolor=blue,
    linkcolor=blue,
    urlcolor=blue
}

\journal{Int J Radiation Oncol Biol Phys}

\begin{document}

\begin{frontmatter}



\title{A learning-driven automatic planning framework for proton PBS treatments of H\&N cancers}

%
\author[1]{Qingqing Wang}
\ead{qiw058@health.ucsd.edu}
\author[2]{Liqiang Xiao}
\ead{liqiangx@amazon.com}
\author[1,3]{Chang Chang}
 \ead{chang2@health.ucsd.edu}
\affiliation[1]{organization={Department of Radiation Medicine and Applied Sciences, University of California at San Diego},
             addressline={9500 Gilman Drive},
             city={La Jolla},
             postcode={92093},
             state={California},
             country={USA}}
\affiliation[2]{organization={Amazon Inc.},
             country={USA}}
 \affiliation[3]{organization={California Protons Cancer Therapy Center},
             addressline={9730 Summers Ridge Road},
             city={San Diego},
             postcode={92121},
             state={California},
             country={USA}}

\begin{abstract}
\noindent \textbf{Purpose:}  
Proton pencil beam scanning (PBS) treatment planning for head \& neck (H\&N) cancers involves numerous conflicting objectives, requiring iterative objective parameter adjustments to balance multiple clinical goals. 
We propose a learning-driven inverse optimizer and integrate it into a proximal policy optimization (PPO)-based planning framework to automatically generate high-quality plans for patients with diverse treatment requirements.\\
\noindent \textbf{Methods and Materials:} 
The inverse optimizer is a learning-to-optimize (L2O) method that predicts update steps by learning from task-specific data distributions. 
For the first time, long-context processing techniques developed for large language models (LLMs) are utilized to address the scalability limitations of existing L2O methods, enabling simultaneous optimization over a substantially large set of variables. 
The PPO framework functions as an outer-loop virtual planner, autonomously adjusting objective parameters through a policy network, and the inner-loop L2O inverse optimizer computes machine-deliverable spot monitor unit (MU) values based on the PPO-refined objectives. Moreover, a Swin~UnetR dose predictor is trained with prescription- and beam-specific information to estimate the initial objective parameters.
In our experiments, total 97 patients with bilateral or ipsilateral H\&N cancers are collected for training and testing. \\
\noindent \textbf{Results:} 
Compared with the second-order gradient-based methods, our L2O optimizer improves the effectiveness and efficiency of the time-consuming inverse optimization by 22.97\% and 36.41\%, respectively, and in conjunction with the PPO-based virtual planner, plans are generated within clinically acceptable times, i.e. 2.55 hours in average, and shows improved or comparable organs-at-risk sparing with superior target coverage compared with human-generated plans.\\
\noindent \textbf{Conclusions:} 
The proposed inverse optimizer is the first L2O model applied to radiotherapy treatment planning and achieves promising performance. The high-quality plans generated for patients with variable prescription dose levels, multiple target volumes and patient-specific beam angles highlight the strong potential of the proposed automatic planning framework for practical clinical use.
\end{abstract}


%
%


\end{frontmatter}



\section{INTRODUCTION}
\label{sec1}

Proton therapy has gained increasing popularity in recent years because of its ability to deliver highly conformal dose distributions, spare surrounding healthy tissues and reduce the risk of radiation-induced side effects. 
These characteristics are particularly advantageous for complex treatment sites such as H\&N cancers, where a large number of organs-at-risk (OARs) are involved, often resulting in conflicting objectives.
These objectives must be carefully balanced to produce dose distributions that meet multiple clinical goals and constraints. 
However, generating high-quality plans that ensure adequate target coverage while minimizing OAR sparing remains both time-consuming and labor-intensive. 
This process typically requires human planners to continuously interact with the treatment planning system (TPS), iteratively evaluating inverse optimization results and adjusting objective parameters. 
The quality of generated plans therefore depends on the planners' experience, the system's optimization performance and the time available for planning. 
In practice, plans generated by expert human planners may still fail to meet all clinical criteria due to limited planning time and the inherent performance constraints of the TPS, often necessitating trade-offs between competing objectives. 
These challenges underscore the urgent need for automatic treatment planning systems that are clinically practical and capable of consistently producing high-quality treatment plans.

Recent studies on automatic treatment planning focus on automatically adjusting objective parameters using deep reinforcement learning (DRL)~\cite{A01, A03, A04, A06, A12}. 
For example, Shen et al.~\cite{A01,A03} employed Q-learning~\cite{A05} to learn adjustment intelligence in a discrete space for high-dose-rate brachytherapy and prostate cancer intensity-modulated radiation therapy (IMRT).
Yang et al.~\cite{A12} utilized soft actor-critic algorithm to adjust key objective parameters for H\&N cancer IMRT.  
Although these works present promising solutions to enable a learning agent with parameter adjustment intelligence, several limitations hinder their clinical application: 
1) Scalability: network sizes increase linearly with the number of adjustable parameters, limiting their capacity to a small number of objectives; 
2) Flexibility: actions are confined to a small discrete space, resulting in coarse and fixed adjustment steps;
3) Generalizability: patient cohorts are typically restricted to specific prescription dose levels and target volume counts, failing to generalize across diverse clinical scenarios.
To address these challenges, Wang et al.~\cite{A04} proposed the first PPO-based proton PBS treatment planning framework capable of handling a large number of objectives without being constrained by immutable plan settings.
Their policy network operated in a continuous action space and achieved human-level performance in H\&N cancers using adjustable, clinically realistic plan settings.
However, this model was susceptible to prolonged optimization times. 
This is because in proton PBS treatment planning, the most widely-used inverse optimizers usually leverage Hessian matrix approximations, i.e., second-order gradients, to compute optimal search directions and step sizes for superior optimization outcomes. 
Computation cost  increases superlinearly with the number of iterations, becoming markedly expensive when iteration counts exceed 100~\cite{A33}. 
Therefore, the development of more effective and computationally efficient optimization strategies is imperative to further advance clinical practicality.

The emerging L2O paradigm, which learns optimization update strategies from task-specific data distributions, provides a potential solution. Existing works~\cite{A15, A19}, especially those in the meta-learning branch~\cite{A20, A21, A22, A23}, have demonstrated strong empirical performance, achieving substantially faster convergence and requiring an order of magnitude fewer iterations than traditional optimizers on unseen problems. Unfortunately, most existing L2O approaches rely on shallow neural network architectures and encounter scalability challenges due to the considerable memory overhead from unrolled computational graphs, which restricts their use to small-scale problems. 
Although recent studies have introduced Transformer-based L2O models with enhanced representational capacity and scalability~\cite{A22, A23}, their application remains largely confined to neural network parameter optimization, where the underlying architecture is exploited for more efficient learning. 
In addition, these models have shown performance improvements only over first-order optimizers such as Adam~\cite{A17}, and their advantages over the second-order optimizers have so far been demonstrated exclusively on small-scale problems involving fewer than 1,000 independent coordinates~\cite{A19}.

Here we present the first Transformer-based L2O optimizer specifically designed for the direct optimization of proton spots in PBS treatment planning.
Inspired by recent advancements in LLMs that extend context windows, our method incorporates long-context processing techniques into a Transformer encoder to enable simultaneous optimization across a markedly large set of variables, resulting in a computationally efficient and highly scalable optimization process.
Experimental results demonstrate that our L2O-based inverse optimizer substantially improves both effectiveness and computational efficiency compared to the widely-used second-order gradient-based method. 
Most importantly, when integrated with our dose prediction model and PPO-based virtual planner, the automatic planning system achieves human-level plan quality within clinically acceptable planning times.

\section{MATERIALS AND METHODS}
\label{method}

\subsection{Patient cohorts}
Approval from the institutional review board of University of California San Diego was received for this study.
In our experiments, a total of 97 patients with bilateral or ipsilateral H\&N cancers treated at the California Protons Cancer Therapy Center between 2017 and 2024 were included, and the cohort was randomly divided into a training set with 72 patients and a test set with 25 patients.
The number of treatment fractions ranges from 5 to 60 {\it bid}, with up to 5 treatment beams used per plan. 
Each plan included one to four prescription levels, with corresponding prescription doses ranging from 4.0~Gy to 74.2~Gy. 
To better spare OARs, field-specific targets are defined to constrain spot placement during planning.
In our experiments, prescription doses, beam angles, field-specific targets, and target volumes are kept identical to the corresponding clinical plans to enable a fair comparison. 
Hexagonal spot grids are generated around the isocenter with a spot spacing of 7~mm and a layer spacing of 7~mm in water-equivalent thickness (WET). 
Dose calculations are performed using Monte Carlo simulation with a 2~mm scoring voxel spacing. 
To ensure machine deliverability, the proton beam energy is constrained to 70--240~MeV, and the monitor units (MUs) per spot are limited to 3--300~MU.

\subsection{Automatic proton PBS treatment planning framework}

\begin{figure}[h]
\centering
\includegraphics[width=1.0\textwidth]{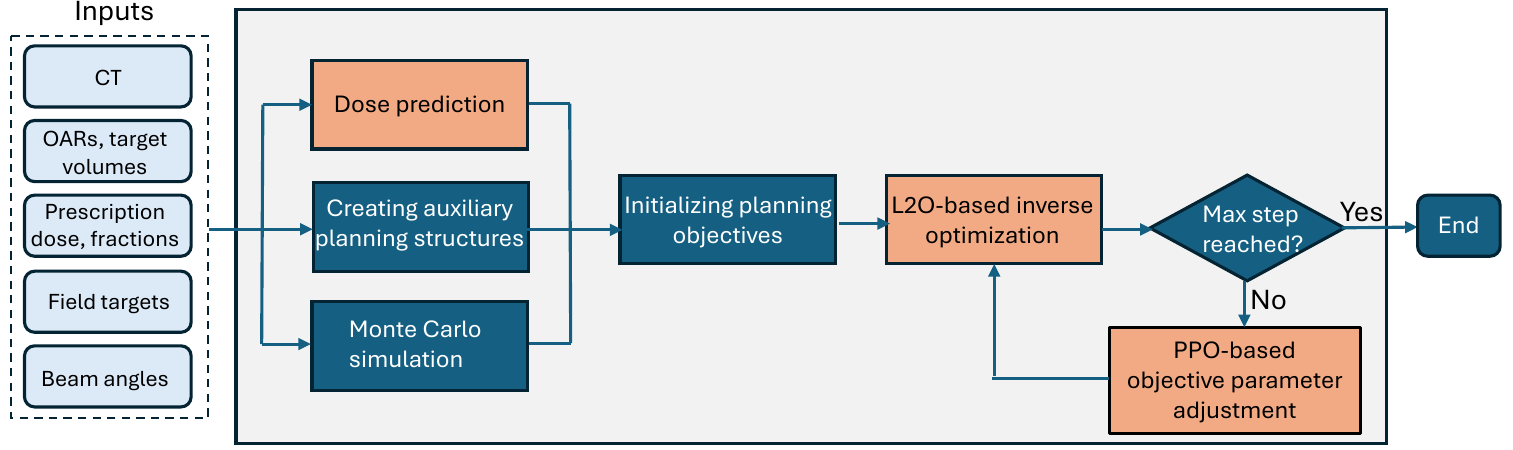}
\caption{Framework of the proposed automatic proton PBS treatment planning method. }
\label{fig1}
\end{figure}

An overview of the proposed automatic proton PBS treatment planning framework is presented in Figure~\ref{fig1}. 
Following the pipeline from Wang's work~\cite{A04}, it employs empirical rules to create auxiliary planning structures and associated objectives, and uses a dose prediction model to estimate initial dose limits for OAR objectives.
Subsequently, an inner-loop inverse optimizer performs plan optimization and an outer-loop PPO-based policy network acts as a virtual planner to iteratively adjust the objective parameters upon the optimization results. 
In this study, the conventional second-order gradient-based Limited-memory Broyden–Fletcher–Goldfarb–Shanno (L-BFGS)~\cite{A37} optimizer is replaced with a novel L2O optimizer. 
The policy network is enhanced with OAR-specific score functions, and the small-scale UnetR dose predictor is upgraded to a large-scale Swin UnetR-based model~\cite{A24} that incorporates prescription- and beam-specific information. 
These improvements are briefly described below.

\begin{figure}[h]
\centering
\includegraphics[width=0.8\textwidth]{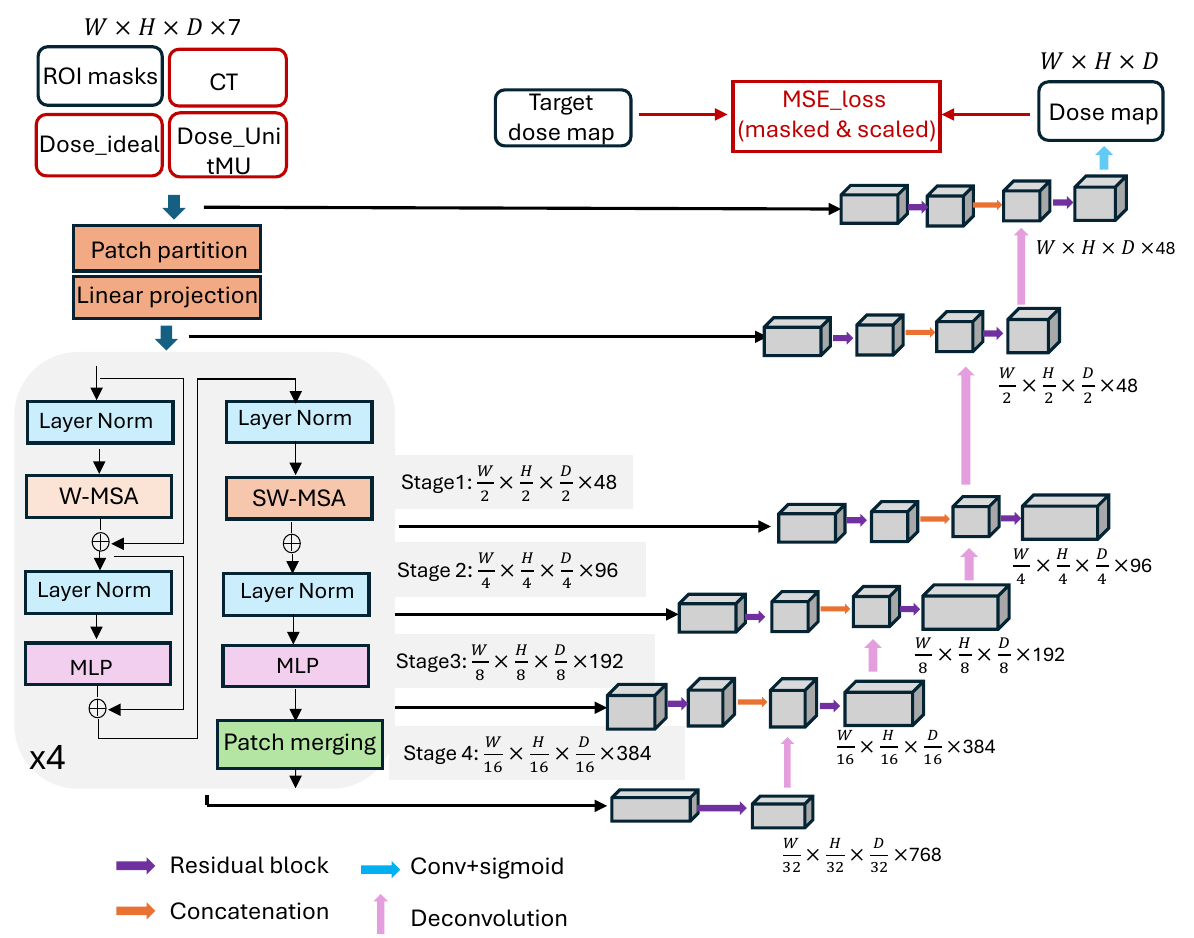}
\caption{Swin UnetR dose prediction model.}
\label{fig2}
\end{figure}

{\bf Dose prediction model.} 
Prior dose predictors are predominantly based on 3D U-Net architectures and take only CT and masks of regions of interest (ROIs) as input~\cite{A27, A34}. 
In contrast, ours is built upon the advanced Swin~UnetR model~\cite{A24}, as depicted in Figure~\ref{fig2}, where in addition to CT and ROI masks, the prescription- and beam-specific information are also taken as input to help better predict patient-specific dose distributions. 
Information such as prescription dose levels, range shifters, and beam angles all inherently influence the achievable planning objectives and the resultant dose distributions. 
Therefore, the dose predictor presented here is more clinically realistic than earlier ones.
Specifically, the Swin UnetR~\cite{A24} backbone is pretrained on 1,470 medical images from the brain tumor segmentation dataset BraTS2021 and fine-tuned on our dose prediction dataset, transferring anatomical prior knowledge from the segmentation task to the dose prediction task.
This enables effective fine-tuning on limited data while mitigating overfitting.
The network input consists of conventional CT images and ROI masks, along with two additional channels, i.e., Dose\_ideal and Dose\_UnitMU, which encode prescription- and beam-specific information, respectively.
Dose\_ideal is generated by assigning the prescribed dose to voxels within the target volumes while setting all other voxels to zero. 
Dose\_UnitMU is computed by summing the rows of the plan's dose influence matrix, representing the dose distribution when the MU values of all spots are set to one. 
In the data preprocessing stage, all input and output volumes are resampled to a uniform size of [256, 256, 128], and normalized to [0, 1].
The ground truth dose distributions obtained from the clinical plans are masked by ROIs and normalized by $1.1\times Rx_{\mathrm{max}}$, where $Rx_{\mathrm{max}}$ denotes the maximum prescription dose among all target volumes, aiming to reduce variability arising from the wide and potentially uneven distribution of clinical prescription dose levels.

{\bf Policy network.} 
The policy network used to adjust objective parameters is trained using the PPO framework guided by a dose distribution-based reward function~\cite{A04}.
This reward function evaluates changes in plan quality, indicated by a plan score, after each adjustment iteration. 
The overall plan score is computed as the sum of individual planning structure scores, each calculated from the structure’s dose distributions.
For each structure, a penalty is applied when voxel doses fall outside a predefined dose interval. 
The dose interval for a CTV with prescription dose $Rx$ was set to $[1.0\times Rx, 1.05\times Rx_{\mathrm{max}}]$. 
For OARs, a simplified fixed interval of [0, 10]~Gy was used earlier, with penalties applied when doses exceeded 10~Gy.
Recognizing that OARs have varying clinical tolerances, here we improved the reward function by introducing individualized, OAR-specific allowable dose intervals as listed in ~\ref{app0:table}.
In addition, actual dose limits used to initialize the optimization are determined by taking the minimum of two values: the clinically derived threshold $D_{clinic}$ from ~\ref{app0:table} and the patient-specific prediction $D_{predict}$ obtained from the dose prediction model described above, i.e., $\min(D_{predict}, D_{clinic})$.
This combined approach ensures that appropriate, patient-tailored dose constraints are used.

\subsection{Learning-to-optimize for inverse optimization}

In this study, the objective function $f(x)$ for proton PBS treatment planning is defined as:
\begin{align} 
\begin{split}
f(X)&=  \sum_{\;i\in \mathcal{S}_{D_{max}}} \!\!\!w^i \biggl(\frac{1}{N^i}\sum\bigl\lVert M^iX - D_{max}^i \bigr\rVert^2_+ \biggr) \\
&\quad + \sum_{j \in \mathcal{S}_{D_{min}}} \!\!\!w^j \biggl( \frac{1}{N^j} \sum\bigl\lVert M^j X - D_{min}^j \bigr\rVert^2_- \biggr)  \\
&\quad + \sum_{k\in \mathcal{S}_{D_{mean}}} \!\!\!w^k\biggl( \Bigl\lVert\frac{1}{N^k}\sum M^k X - D_{mean}^k \Bigr\rVert^2_+ \biggr)  ,
\label{eq:eq1}
\end{split}
\end{align}
and the goal is to achieve the optimal spot MU values $X^{^{\ast}}$ which minimizes $f(X)$, i.e., $X^{^{\ast}}=\argmin_{X}\,f(X)$. 
Here, $\mathcal{S}_{D_{max}}$, $\mathcal{S}_{D_{min}}$ and $\mathcal{S}_{D_{mean}}$ represent sets of planning structures with objectives on maximum dose, minimum dose, and mean dose, respectively. 
The terms $w^i$, $w^j$ and $w^k$ denote the corresponding weights, and $D_{max}^i$, $D_{min}^j$ and $D_{mean}^k$ represent the associated dose limits for these objectives. 
For planning structures $i$, $j$, $k$ from $\mathcal{S}_{D_{max}}$, $\mathcal{S}_{D_{min}}$ and $\mathcal{S}_{D_{mean}}$,  $N^i$, $N^j$ and $N^k$ denote the numbers of voxels within each structure, and $M^i$, $M^j$ and $M^k$ denote the corresponding dose influence matrices. 
Notations $\lVert\cdot\rVert^2_+$ and $\Vert\cdot\rVert^2_-$ are the standard $l_2$-norms that compute only the positive and the negative elements, respectively.
Objective parameters $w^i$, $w^j$, $w^k$, $D_{max}^i$, $D_{min}^j$ and $D_{mean}^k$ are adjusted by the policy network in the outer loop, but they are set to fixed values during the inverse optimization stage in the inner loop.

The widely-used second-order gradient-based optimizers, such as L-BFGS~\cite{A37}, update $X$ with $X_{t+1}=X_t-\varphi H_t \, g_t$ at time step $t+1$, where $g_t$ and $H_t$ are the gradient and the inverse Hessian of $X_t$, respectively. 
The update step $\varphi$ is typically determined using strategies such as line search. 
In contrast, L2O updates $X$ with $X_{t+1}=X_t-\mathcal{N}(Z_t, \phi)$, where $\mathcal{N}(\cdot,\cdot)$ is an update rule learned by a neural network parameterized with $\phi$, and $Z_t$ denotes the input of the network.
The $\phi$-parameterized network is trained by minimizing the loss function $L(\phi)$:
\begin{equation}
L(\phi) = \mathbb{E}_{f\in \Gamma}\Biggl[\sum_{t=1}^T w_t f(X_t)\Biggr]
\label{eq:eq2}
\end{equation}
where $\Gamma$ is an ensemble of optimization problems or tasks to be solved. In our task, $\Gamma$ is the ensemble of patients and $T$ is the time span, i.e. unrolling length, over which the $w_t$-weighted sum of objectives are formulated as the loss of network prediction.

\begin{figure}[h]
\centering
\includegraphics[width=1.0\textwidth]{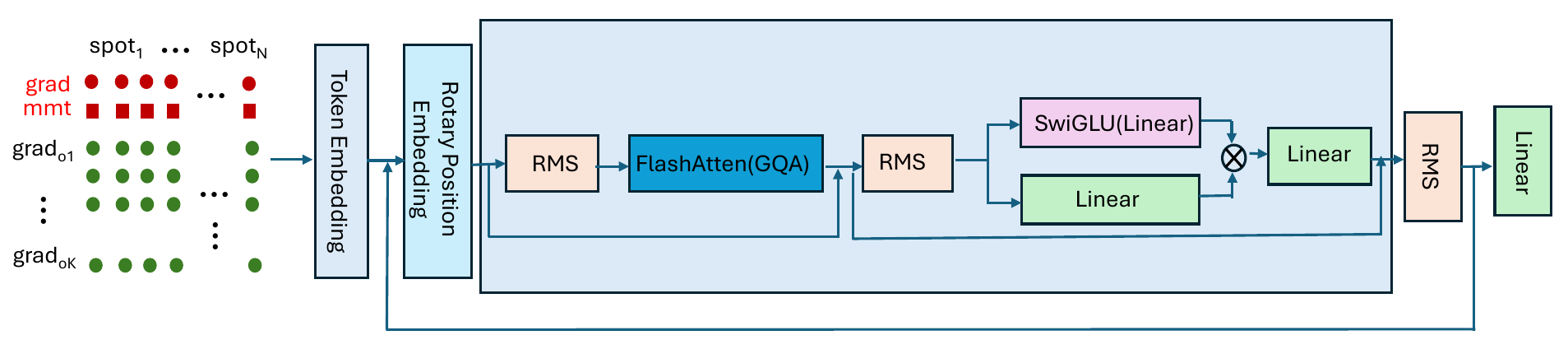}
\caption{Structure of the proposed learning-to-optimize model. }
\label{fig3}
\end{figure}

The framework of our L2O network is presented in Figure~\ref{fig3}. 
It is a 6-layer Transformer encoder equipped with the following techniques for long-context representational capacity:
\begin{enumerate}
	\item Group query attention (GQA)~\cite{A30} used to reduce GPU memory usage when calculating attentions;
	\item Rotary positional embeddings (RoPE)~\cite{A29} with a base frequency hyper-parameter of 500,000 used to better support longer input sequences;
	\item RMSNorm mormalization~\cite{A31} and SwiGLU activation function used to improve the training stability and the performance;
	\item FLASHATTENTION~\cite{A32} used to speed up the training by reducing the number of memory reads/writes between the GPU's high bandwidth memory and its on-chip SRAM.
\end{enumerate}
These techniques are inherited from LLMs like Llama~\cite{A25}, with the distinction that LLMs are decoder models predicting output tokens sequentially, while our network is an encoder model that predicts spot MU values simultaneously.

This L2O network takes gradients and momentums computed from the overall objective function defined in Equation~(\ref{eq:eq1}) as input. 
We also find that more detailed information, particularly, how each spot contributes to individual components of the objective function, can assist the optimizer in identifying more effective update steps.
Accordingly, the network input is composed of three components: the overall gradient, the overall momentum, and the split gradients corresponding to each of the planning objectives. These components are normalized using a strategy similar to that of Adam~\cite{A17} (see~\ref{app1:norm} for details).

In practice, treatment machines impose constraints on spot MU values. 
Therefore, to ensure the generation of machine-deliverable plans, our L2O optimizer limits the per-spot MU values to the range of 3--300~MU after the first 50 optimization iterations and removes spots with extremely small MU values.
In addition, network training can become unstable as the model approaches convergence, potentially due to significant variations in loss values across updates. To address this issue, the loss function defined in Equation~(\ref{eq:eq1}) is modified following \ref{app2:loss} in our experiments.

\section{RESULTS}
\subsection{Implementation details}
The L2O network is trained using the AdamW optimizer~\cite{A38} with a learning rate of 0.0004. 
The virtual planner manages up to 32 OARs and 76 adjustable objective parameters, and the L2O optimizer simultaneously predicts update steps for up to 25,000 spots. Clinical reference plans, denoted as Plan\_human, are manually created by experienced dosimetrists and have been approved by radiation oncologists for patient treatment.
Experiments are conducted on a workstation equipped with two NVIDIA RTX 6000 Ada GPUs and a Ryzen Threadripper PRO 5975WX CPU.

\subsection{Evaluations}
\subsubsection{Performance evaluation of the L2O optimizer}
We assess the effectiveness and efficiency of the proposed L2O method by comparing its performance to that of L-BFGS-B~\cite{A36}, a quasi-Newton optimization algorithm that extends the L-BFGS by supporting box constraints, i.e., simple upper and lower bounds, on the variables being optimized.   

\setlength{\tabcolsep}{2pt}
\begin{figure}[htbp]
\scriptsize
\centering
\begin{minipage}[t]{0.48\textwidth}
\centering
\rotatebox{90}{
\begin{minipage}{\textheight}
\centering
\captionof{table}{Effectiveness and efficiency improvements of proposed L2O optimizer on patients with bilateral and ipsilateral H\&N cancers.}
\begin{tabular}{|c|l|l|l|l|l|l|l|l|l|l|l|l|l|l|l|l|l|}
\hline
\multicolumn{1}{|l|}{}                                                                                          &                                                                                & B03     & B08     & B12     & B15     & B17     & B19     & B24     & B27     & B30     & B30\_v2 & B35    & B38     & B38\_bst & B41     & B43     & overall     \\ \hline
\multirow{3}{*}{L-BFGS-B}                                                                                       & iterations                                                                     & 100     & 100     & 100     & 100     & 100     & 100     & 100     & 100     & 100     & 100     & 100    & 100     & 100      & 100     & 100     & \textbackslash{} \\ \cline{2-18} 
                                                                                                                & \multicolumn{1}{c|}{\begin{tabular}[c]{@{}c@{}}time \\ (seconds)\end{tabular}} & 4719.17 & 1436.69 & 1119.06 & 1993.17 & 1359.77 & 587.05  & 783.28  & 1787.00 & 3470.15 & 3865.72 & 468.36 & 2626.04 & 1019.48  & 1877.28 & 1474.05 & \textbackslash{} \\ \cline{2-18} 
                                                                                                                & \multicolumn{1}{c|}{loss\_min}                                                 & 22.58   & 1.87    & 4.87    & 9.17    & 4.51    & 6.70    & 3.95    & 1.14    & 36.20   & 41.96   & 3.61   & 1.58    & 0.22     & 13.77   & 1.86    & \textbackslash{}  \\ \hline
\multirow{4}{*}{\begin{tabular}[c]{@{}c@{}}L2O\\(within the\\same\\optimization\\time)\end{tabular}}  & iterations                                                                     & 164     & 144     & 196     & 185     & 186     & 161     & 127     & 141     & 133     & 130     & 132    & 138     & 167      & 149     & 160     & \textbackslash{} \\ \cline{2-18} 
                                                                                                                & \multicolumn{1}{c|}{\begin{tabular}[c]{@{}c@{}}time\\ (seconds)\end{tabular}}  & 4701.06 & 1427.03 & 1118.28 & 1984.09 & 1354.73 & 583.86  & 780.26  & 1785.68 & 3452.63 & 3860.78 & 467.21 & 2618.54 & 1017.78  & 1868.80 & 1465.03 & \textbackslash{} \\ \cline{2-18} 
                                                                                                                & loss\_min                                                                      & 16.35   & 0.85    & 3.95    & 8.24    & 1.72    & 5.97    & 3.98    & 0.39    & 31.57   & 36.95   & 3.46   & 0.67    & 0.15     & 9.51    & 1.04    & \textbackslash{} \\ \cline{2-18} 
                                                                                                                & \begin{tabular}[c]{@{}l@{}}effectiveness$\uparrow$\end{tabular}                   & 27.59\% & 54.55\% & 18.89\% & 10.14\% & 61.86\% & 10.90\% & -0.76\% & 65.79\% & 12.79\% & 11.94\% & 4.16\% & 57.59\% & 31.82\%  & 30.94\% & 44.09\% & 29.49\%      \\ \hline
\multirow{4}{*}{\begin{tabular}[c]{@{}c@{}}L2O\\(achieve the\\same\\loss\_min)\end{tabular}} & iterations                                                                     & 55      & 66      & 91      & 85      & 69      & 112     & 132     & 66      & 63      & 61      & 122    & 59      & 82       & 67      & 65      & \textbackslash{} \\ \cline{2-18} 
                                                                                                                & \multicolumn{1}{c|}{\begin{tabular}[c]{@{}c@{}}time\\ (seconds)\end{tabular}}  & 1576.58 & 654.06  & 519.20  & 911.61  & 502.56  & 406.16  & 810.98  & 835.85  & 1635.46 & 1811.59 & 431.81 & 1119.52 & 499.75   & 840.33  & 595.17  & \textbackslash{}\\ \cline{2-18} 
                                                                                                                & loss\_min                                                                      & 22.58   & 1.86    & 4.84    & 9.15    & 4.50    & 6.69    & 3.95    & 1.11    & 36.01   & 41.96   & 3.60   & 1.56    & 0.21     & 13.57   & 1.84    & \textbackslash{} \\ \cline{2-18} 
                                                                                                                & \begin{tabular}[c]{@{}l@{}}efficiency$\uparrow$\end{tabular}                       & 66.59\% & 54.47\% & 53.62\% & 54.26\% & 63.04\% & 30.81\% & -3.54\% & 53.23\% & 52.87\% & 53.14\% & 7.80\% & 57.37\% & 50.98\%  & 55.24\% & 59.62\% & 47.30\%      \\ \hline
\end{tabular}
\label{tab:tab2}
\end{minipage}
}
\end{minipage}
\hfill
\begin{minipage}[t]{0.48\textwidth}
\centering
\rotatebox{90}{
\begin{minipage}{\textheight}
\centering
\begin{tabular}{|c|l|l|l|l|l|l|l|l|l|l|l|l|}
\hline
\multicolumn{1}{|l|}{}                                                                                          &                                                                                &  I06      & I11     & I15     & I19     & I28     & I30      & I37     & I40     & I43     & I46     & overall        \\ \hline
\multirow{3}{*}{L-BFGS-B}                                                                                       & iterations                                                                     & 100      & 100     & 100     & 100     & 100     & 100      & 100     & 100     & 100     & 100     &  \textbackslash{} \\ \cline{2-13} 
                                                                                                                & \multicolumn{1}{c|}{\begin{tabular}[c]{@{}c@{}}time \\ (seconds)\end{tabular}} & 558.12   & 866.46  & 618.98  & 539.63  & 163.78  & 687.05   & 498.98  & 284.54  & 511.55  & 482.64  & \textbackslash{}  \\ \cline{2-13} 
                                                                                                                & \multicolumn{1}{c|}{loss\_min}                                                 & 5.66     & 1.72    & 4.03    & 3.95    & 1.65    & 7.47     & 5.19    & 4.60    & 3.84    & 11.01   & \textbackslash{}  \\ \hline
\multirow{4}{*}{\begin{tabular}[c]{@{}c@{}}L2O\\(within the\\same\\optimization\\time)\end{tabular}}  & iterations                             & 155      & 163     & 140     & 176     & 185     & 156      & 169     & 194     & 138     & 136     & \textbackslash{}  \\ \cline{2-13} 
                                                                                                                & \multicolumn{1}{c|}{\begin{tabular}[c]{@{}c@{}}time\\ (seconds)\end{tabular}}  & 555.62   & 864.58  & 618.61  & 537.44  & 163.39  & 684.30   & 498.82  & 284.02  & 511.41  & 479.82  & \textbackslash{}  \\ \cline{2-13} 
                                                                                                                & loss\_min                                                                       & 5.84     & 1.42    & 3.70    & 3.69    & 1.43    & 7.68     & 4.21    & 4.47    & 3.09    & 5.37    & \textbackslash{}  \\ \cline{2-13} 
                                                                                                                & \begin{tabular}[c]{@{}l@{}}effectiveness$\uparrow$\end{tabular}            & -3.18\%  & 17.44\% & 8.19\%  & 6.58\%  & 13.33\% & -2.81\%  & 18.88\% & 2.83\%  & 19.53\% & 51.23\% & 13.20\%          \\ \hline
\multirow{4}{*}{\begin{tabular}[c]{@{}c@{}}L2O\\(achieve the\\same\\loss\_min)\end{tabular}} & iterations                                           & 181      & 98      & 115     & 137     & 133     & 185      & 100     & 165     & 102     & 76      & \textbackslash{}  \\ \cline{2-13} 
                                                                                                                & \multicolumn{1}{c|}{\begin{tabular}[c]{@{}c@{}}time\\ (seconds)\end{tabular}}   & 648.82   & 519.81  & 508.14  & 418.35  & 117.47  &  811.50    & 295.16  & 241.56  & 378.00  & 268.14  & \textbackslash{}  \\ \cline{2-13} 
                                                                                                                & loss\_min                                                                     & 5.65     & 1.72    & 4.02    & 3.95    & 1.65    &7.46  & 5.17    & 4.60    & 3.82    & 10.42   & \textbackslash{}  \\ \cline{2-13} 
                                                                                                                & \begin{tabular}[c]{@{}l@{}}efficiency$\uparrow$\end{tabular}                       & -16.25\% & 40.01\% & 17.91\% & 22.47\% & 28.28\% & -18.11\% & 40.85\% & 15.11\% & 26.11\% & 44.44\% & 20.08\%             \\ \hline
\end{tabular}
\end{minipage}
}
\end{minipage}
\end{figure}

For each patient, L-BFGS-B and L2O are compared using two criteria: 
(1) minimal loss achieved within the time required for L-BFGS-B to complete 100 iterations, and 
(2) time required to reach the loss level obtained by L-BFGS-B at 100 iterations. 
These criteria measure effectiveness and efficiency, respectively.
For example, in case B03, L-BFGS-B takes 4719.17 seconds for 100 iterations to reduce the objective loss to 22.58.
By contrast, the proposed L2O algorithm can reduce the loss to a lower value of 16.35 within a slightly shorter optimization time of 4701.06 seconds, resulting in an effectiveness improvement of 27.59\%. 
Similarly, to reach the same minimal loss of 22.58, the L2O algorithm requires only 1576.58 seconds, leading to an efficiency improvement of 66.59\%.
Table~\ref{tab:tab2} demonstrates the detailed effectiveness and efficiency gains for each patient.

On average, the proposed L2O algorithm improves optimization effectiveness and efficiency by 22.97\% and 36.41\%, respectively, across all 25 test patients, including 15 with bilateral H\&N cancers and 10 with ipsilateral H\&N cancers. 
Improvements in both effectiveness and efficiency are observed in 22 out of 25 patients (88\%), with particularly notable gains in the more complex and time-consuming bilateral cases, where effectiveness and efficiency are improved by 29.49\% and 47.3\%, respectively, compared with the values of 13.20\% and 20.08\% of ipsilateral cases.

 \begin{figure}[htbp]
  \centering
  \begin{subfigure}{0.95\textwidth}
  \centering
   \includegraphics[width=\textwidth]{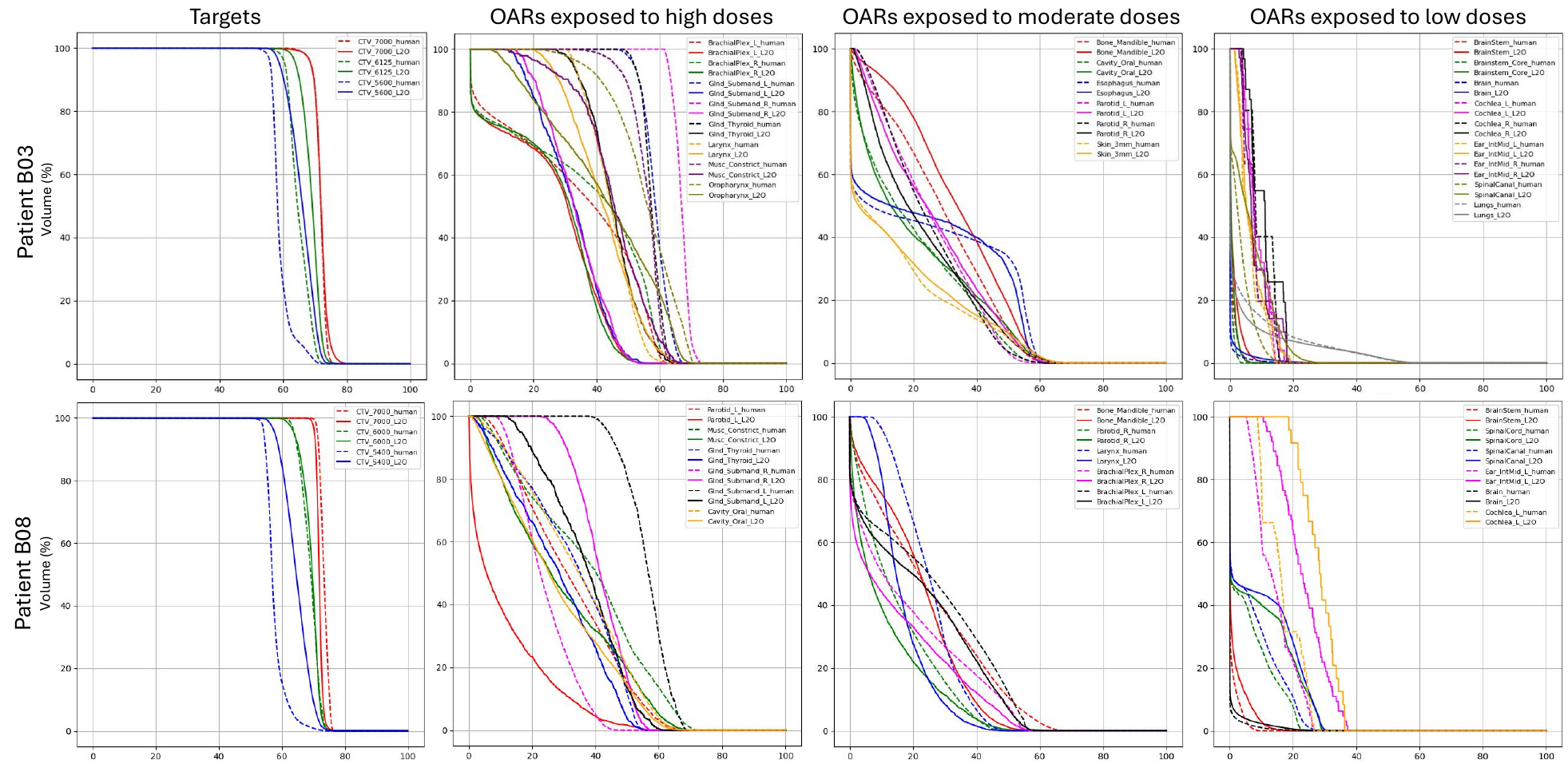}
  \end{subfigure}
  \begin{subfigure}{0.95\textwidth}
  \centering
    \includegraphics[width=\textwidth]{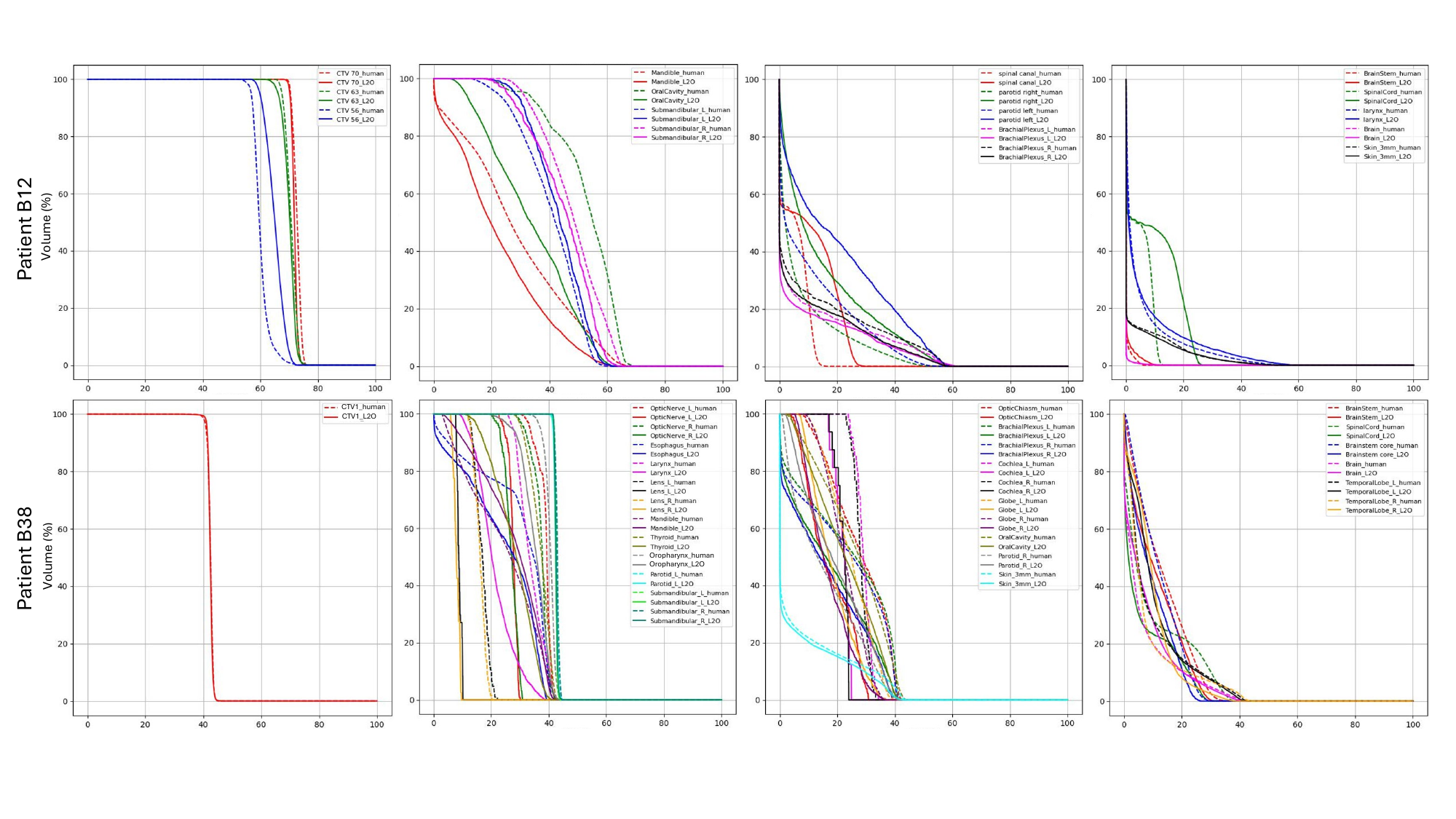}
  \end{subfigure}
   \begin{subfigure}{0.95\textwidth}
   \centering
    \includegraphics[width=\textwidth]{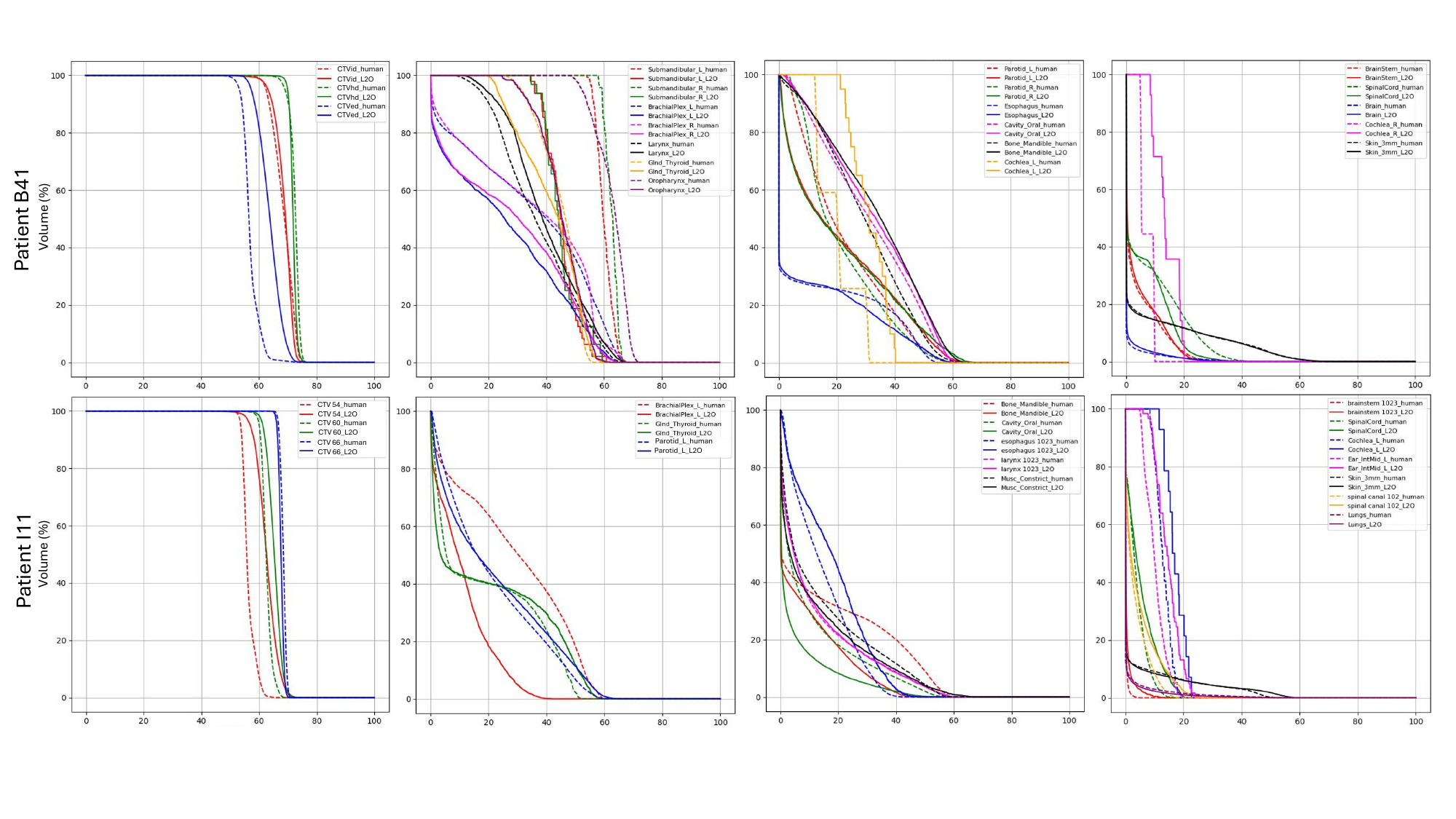}
  \end{subfigure}
  \end{figure}
  \clearpage
  \begin{figure}[htbp]
  \ContinuedFloat
  \centering
   \begin{subfigure}{0.95\textwidth}
   \centering
    \includegraphics[width=\textwidth]{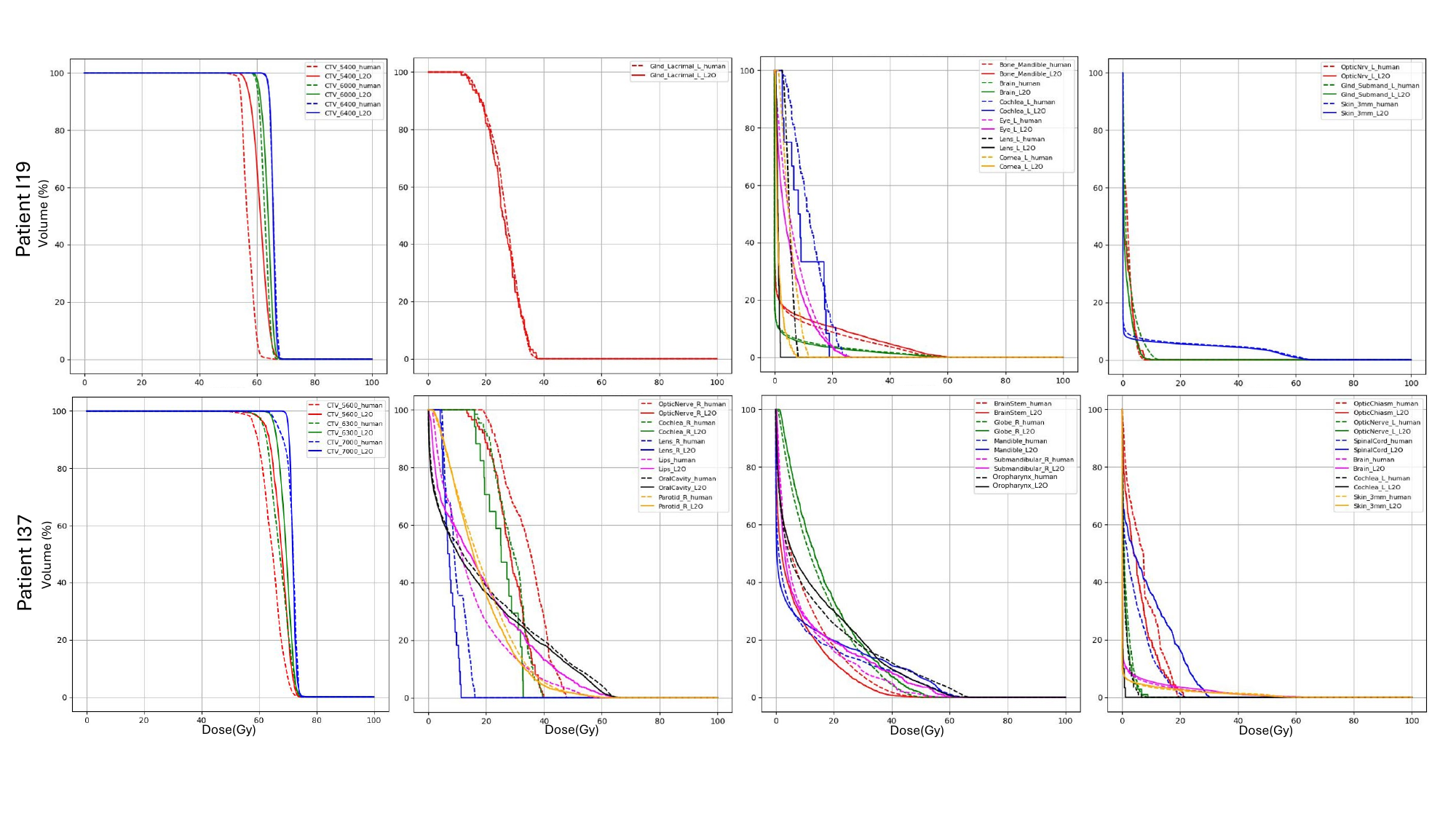}
  \end{subfigure}
    \begin{subfigure}{0.95\textwidth}
   \centering
    \includegraphics[width=\textwidth]{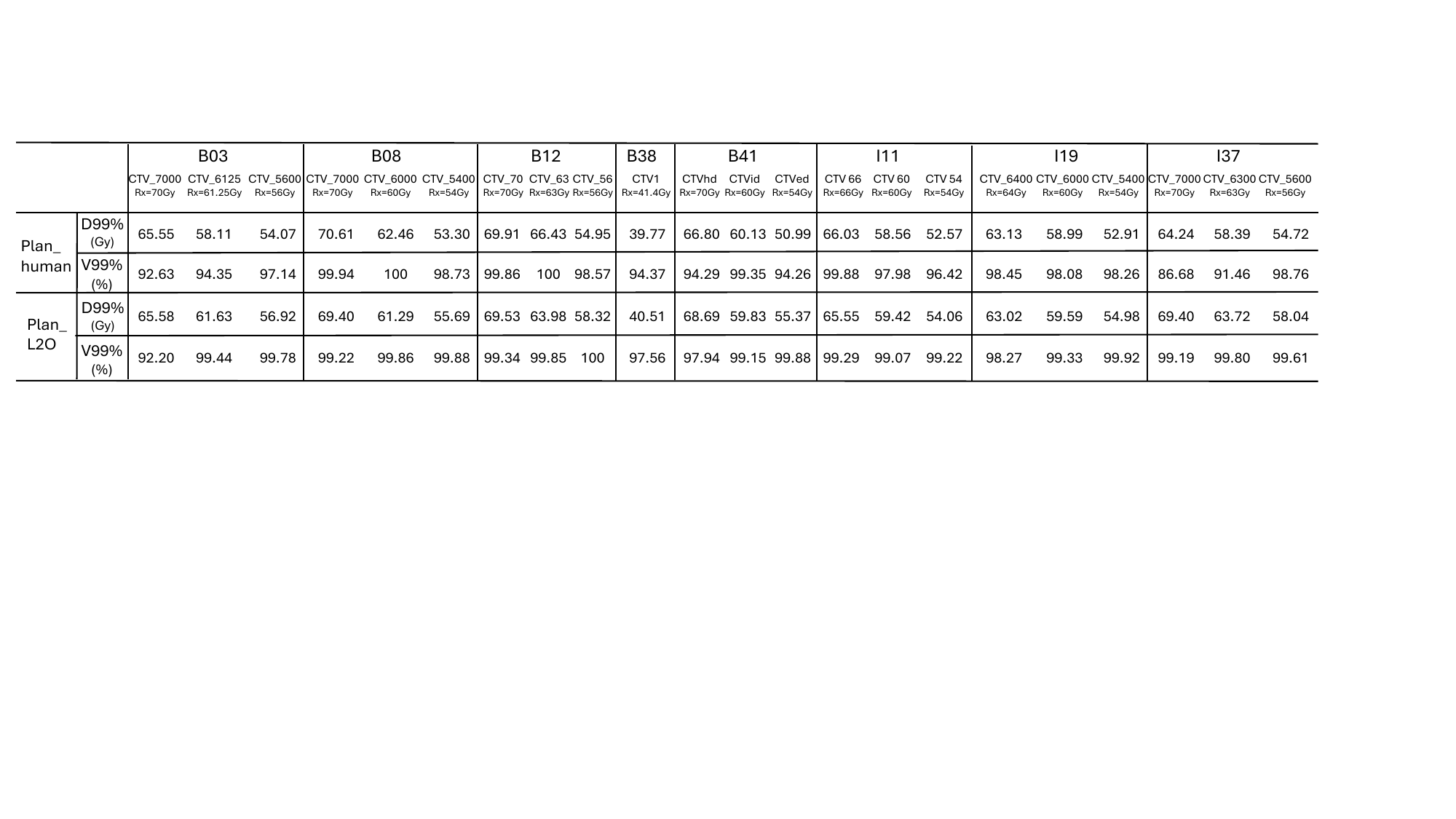}
  \end{subfigure}
  \caption{Comparison results between plans generated by our learning-based automatic planning framework (Plan\_L2O, solid lines) and human-generated plans (Plan\_human, dashed lines).}
  \label{fig4}
\end{figure}

\subsubsection{Comparison with human-generated plans}
\label{exp:com}

We compare the quality of plans generated by our learning-driven automatic planning framework (Plan\_L2O) with that of human-generated clinical plans (Plan\_human). 
For each patient, the policy network adjusts objective parameters for four times, resulting in five inverse optimization steps in total. 
To ensure clinically acceptable runtimes, each inverse optimization is limited to 200 iterations.
On average, the computational time required to generate five plans per patient is 2.55 hours for the 25 test patients. 
Figure~\ref{fig4} presents paired comparisons of Plan\_L2O and Plan\_human for eight representative test patients.
For ease of visualization, OARs are grouped by the dose levels they receive.

For each patient, we select from the five policy-generated plans the one whose target coverage just surpasses that of the human plan; if none do, the plan with the highest coverage is used for comparison.
Using $D_{99\%}$ and $V_{99\%}$ as the criteria for target coverages, detailed numbers are given in Figure~\ref{fig4}. 
Specifically, in case B03, Plan\_L2O achieves higher $D_{99\%}$ than Plan\_human on all of its three target volumes, and despite the slightly lower $V_{99\%}$ of CTV\_7000 in Plan\_L2O than in Plan\_human (92.20\% vs. 92.63\%), Plan\_L2O significantly outperforms Plan\_human on the other two CTVs  in $V_{99\%}$, with 99.44\% vs. 94.35\% and 99.78\% vs. 97.14\%, respectively.
For case B08 and case B12, Plan\_L2O satisfies the clinical requirements of $D_{99\%}>99\%Rx$ and $V_{99\%}>99\%$ on all of their target volumes, but Plan\_human fails the CTV\_5400 of case B08 and the CTV 56 of case B12 with $D_{99\%}$ of 53.30~Gy and 54.95~Gy, and $V_{99\%}$ of 98.73\% and 98.57\%, respectively.
Case B38 is challenging due to tumor's anatomy, completely encompassing three OARs and being in the proximity of 19 other OARs. Therefore, both plans suffer from insufficient target coverage on its only target volume. Nevertheless, Plan\_L2O still outperforms Plan\_human by large margins, with $D_{99\%}$ of 40.51 vs. 39.77~Gy, and $V_{99\%}$ of 97.56\% vs. 94.37\%, respectively.
Case B41 is also a difficult case, where Plan\_human fails two of its three CTVs,  i.e., CTVhd and CTVed, with $D_{99\%}$ of 66.80~Gy and 50.99~Gy, and $V_{99\%}$ of 94.29\% and 94.26\%, respectively. 
In contrast, Plan\_L2O fails only CTVhd, with significantly improved $D_{99\%}$ of 68.69~Gy and $V_{99\%}$ of 97.94\%.
For case I11, Plan\_L2O meets clinical criteria on all three target volumes, but Plan\_human fails CTV~60 and CTV~54 with significantly lower $D_{99\%}$ of 58.56~Gy and 52.57~Gy, and $V_{99\%}$ of 97.98\% and 96.42\%.
For case I19, Plan\_L2O outperforms Plan\_human on CTV\_5400 and CTV\_6000 with superior $D_{99\%}$ (54.98 vs. 52.91~Gy, 59.59 vs. 58.99~Gy) and $V_{99\%}$ (99.92\% vs. 98.26\%, 99.33\% vs. 98.08\%), and achieves nearly equivalent performance on CTV\_6400 (63.02 vs. 63.13~Gy in $D_{99\%}$ and 98.27\% vs. 98.45\% in $V_{99\%}$).
Finally, for case I37, Plan\_L2O meets the clinical criteria on all target volumes, but Plan\_human fails all of them with observably inferior performance, i.e., $D_{99\%}$ of 54.72~Gy, 58.39~Gy and 64.24~Gy, and $V_{99\%}$ of 98.76\%, 91.46\% and 86.68\%, respectively.

Plan\_L2O clearly outperforms Plan\_human in target coverage for all of the above patients. 
With this advantage, Plan\_L2O also achieves comparable, or even superior, OAR sparing. 
For example, case B03 has eight OARs exposed to high doses: BrachialPlex\_L\&R, Glnd\_Submand\_L\&R, Glnd\_Thyroid, Larynx, Musc\_Constrict and Oropharynx.
Plan\_L2O noticeably reduces the mean doses for these OARs by 7.6~Gy, 9.07~Gy, 26.51~Gy, 33.92~Gy, 12.16~Gy, 2.46~Gy, 10.52~Gy and 13.06~Gy, respectively, compared to Plan\_human. 
Especially for the $D_{0.03cc}$ of BrachialPlex\_L\&R, the values are reduced to 53.36~Gy and 53.33~Gy in Plan\_L2O from Plan\_human's 62.34~Gy and 63.59~Gy. 
Among OARs receiving moderate doses, Plan\_human outperforms Plan\_L2O on Bone\_Mandible and Parotid\_L, but both plans satisfy the clinical requirements $D_{0.03cc}<=70$~Gy and $V_{60Gy}<=20\%$ for Bone\_Mandible, and Dmean$<26$~Gy for Parotid.
For the remaining OARs including Cavity\_Oral, Esophagus, Parotid\_R and Skin\_3mm, these two plans achieve comparable results, with mean doses below 25~Gy in both plans.
For case B08, eleven OARs are exposed to high or moderate doses, and ten of these eleven OARs receive significantly lower doses in Plan\_L2O than in Plan\_human. 
Particularly, the $D_{0.03cc}$ of Bone\_Mandible is reduced from 67.83~Gy to 59.56~Gy.
Note that while the mean dose of Glnd\_Submand\_L is reduced significantly from 56.10~Gy to 36.77~Gy in Plan\_L2O, and Plan\_L2O also achieves a slightly lower mean dose for overall Glnd\_Submand (39.12 vs. 40.01Gy) when both left and right sides are considered, Glnd\_Submand\_R receives a higher mean dose in Plan\_L2O than in Plan\_human (41.47 vs. 23.91~Gy) due to physician preference. 
For OARs receiving low doses, both plans meet clinical criteria with sufficiently low values.
For case B12, four OARs are exposed to high doses, among which Plan\_human fails OralCavity with an extremely high mean dose of 52.72~Gy and a $V_{35Gy}$ of 91.91\%, given typical clinical requirement $V_{35Gy}<=50\%$. 
In contrast, Plan\_L2O passes OralCavity with a mean dose of 33.59~Gy and a $V_{35Gy}$ of 47.29\%.
Plan\_L2O also achieves lower mean dose of 44.73~Gy on Submandibular\_R and lower $D_{0.03cc}$ of 61.93~Gy on Mandible, compared with the values of 48.00~Gy and 67.64~Gy in Plan\_human, but it is surpassed by Plan\_human on Submandibular\_L with a mean dose of 40.60~Gy, vs. the 43.16~Gy in Plan\_L2O. 
For OARs receiving moderate and low doses, Plan\_human provides better sparing for Parotid\_left\&right, whereas Plan\_L2O better spares BrachialPlexus\_L\&R, The remaining OARs receive marginal doses in both plans. 
Notably, the mean doses of both Parotids remain below 20Gy in Plan\_L2O, i.e., 19.58~Gy and 14.80~Gy.
As mentioned above, case B38 is challenging because its Parotid\_L and Submandibular\_L\&R are completely enclosed by the target volume and the other 19 OARs are also exposed to high or moderate doses.
For the 19 OARs, Plan\_L2O outperforms Plan\_human on 17 of them, with mean dose improvements of 5.18~Gy, 12.55~Gy, 9.06~Gy, 5.26~Gy, 7.36~Gy, 6.38~Gy, 6.19~Gy, 5.79~Gy, 3.21~Gy, 7.86~Gy and 8.10~Gy for Esophagus, Larynx, Thyroid, Oropharynx, Cochlea\_L\&R, Globe\_L\&R, OralCavity,  Lens\_L\&R,
 and $D_{0.03cc}$ improvements of 11.72~Gy, 10.62~Gy, 10.56~Gy, 0.39~Gy, 2.18~Gy and 0.87~Gy for OpticNerve\_L\&R, OpticChiasm, BrachialPlexus\_L\&R and Skin\_3mm, respectively.
For the highly radiation-sensitive OAR Lens, whose clinical requirement is $D_{0.03cc}<=10$~Gy, Plan\_human fails both Lens\_L and Lens\_R with the $D_{0.03cc}$ of 19.54~Gy and 18.63~Gy, which are noticeably reduced to 10.10~Gy and 9.34~Gy in Plan\_L2O. 
On the other hand, Plan\_L2O is outperformed by Plan\_human on Mandible and Parotid\_R, but the margins are narrow and both plans achieve $D_{0.03cc}$ below 70~Gy for Mandible (43.27 vs. 42.35~Gy), and mean doses below 26~Gy for Parotid\_R (19.20 vs. 15.66~Gy), which are clinically acceptable.
For case B41, only seven OARs are exposed to high doses, and Plan\_L2O achieves better OAR sparing on six of them. 
In particular, Submandibular\_L\&R and Oropharynx receive substantial mean doses of 60.01~Gy, 62.51~Gy, and 62.71~Gy, respectively, in Plan\_human. 
These values are significantly reduced to 45.09~Gy, 45.13~Gy, and 45.15~Gy in Plan\_L2O. 
Plan\_L2O also reduces the mean doses to BrachialPlex\_L\&R and Glnd\_Thyroid by 8.78~Gy, 7.05~Gy and 3.32~Gy, respectively.
Moreover, despite the slight increases for Larynx and other six OARs receiving moderate doses in Plan\_L2O, both plans satisfy the associated clinical requirements.

Patients with ipsilateral H\&N cancers typically involve fewer OARs, and are therefore easier to plan. 
For example, case I11 has only three OARs exposed to high doses: BrachialPlex\_L, Glnd\_Thyroid and Parotid\_L, while all other OARs receive mean doses below 20~Gy in both plans. 
Notably, BrachialPlex\_L receives the highest mean dose of 28.69~Gy and a $D_{0.03cc}$ of 59.51~Gy in Plan\_human, which are significantly reduced to 11.07~Gy and 39.66~Gy in Plan\_L2O.
The mean doses for Glnd\_Thyroid (19.26 vs. 18.39~Gy) and Parotid\_L (21.67 vs. 21.07~Gy) are slightly higher in Plan\_L2O than in Plan\_human, though all values meet clinical requirements.
Case I19 has only one OAR of concern: Glnd\_Lacrimal\_L, with mean dose and $D_{0.03cc}$ of 25.97~Gy and 35.39~Gy in Plan\_L2O, compared with the 26.50~Gy and 34.83~Gy in Plan\_human. 
These values are comparable, and given the clinical criteria of $D_{0.03cc}<=40$~Gy and Dmean$<=30$~Gy, both plans provide satisfactory OAR sparing. 
Additionally, although all remaining OARs receive mean doses below 10~Gy in both plans, Bone\_Mandible and Brain exhibit high $D_{0.03cc}$ values: 61.10~Gy and 61.46~Gy in Plan\_L2O, and 59.53~Gy and 62.66~Gy in Plan\_human, respectively. 
These results are clinically acceptable and reflect the trade-off necessary to ensure adequate target coverage; both plans demonstrate comparable performance in this regard.
For case I37, six OARs are exposed to high mean doses. 
Plan\_L2O performs better than Plan\_human on five of them: OpticNerve\_R, Cochlea\_R, Lens\_R, OralCavity and Parotid\_R, with respective mean dose improvements of 5.73~Gy, 3.55~Gy, 2.21~Gy,  1.11~Gy and 0.93~Gy.
Especially, the $D_{0.03cc}$ of OpticNerve\_R is significantly reduced from 46.55~Gy in Plan\_human to 38.44~Gy in Plan\_L2O.
Plan\_human outperforms Plan\_L2O only on Lips, with mean doses of 15.09 vs. 18.32~Gy, both satisfying the clinical requirement of Dmean$<=20$~Gy. 
Furthermore, all five OARs exposed to moderate doses receive mean doses below 20~Gy. 
However, the $D_{0.03cc}$ of BrainStem, Mandible and Oropharynx reach 48.43~Gy, 65.20~Gy, and 67.74~Gy, respectively, in Plan\_human, and are slightly reduced to 46.23~Gy, 64.40~Gy, and 65.59~Gy in Plan\_L2O. 
Conversely, Plan\_human yields lower $D_{0.03cc}$ than Plan\_L2O for Globe\_R (48.39 vs. 53.72~Gy) and Submanibular\_R (52.04 vs. 58.53~Gy). 
Given that Plan\_L2O demonstrates outstanding target coverage, with all three target volumes meeting clinical criteria of $D_{99\%}>99\%$ and $V_{99\%}>99\%$, while the corresponding $V_{99\%}$ values in Plan\_human are 98.76\%, 91.46\% and 86.68\%, Plan\_L2O performs better than Plan\_human overall.

\begin{figure}[h]
\centering
\includegraphics[width=\textwidth]{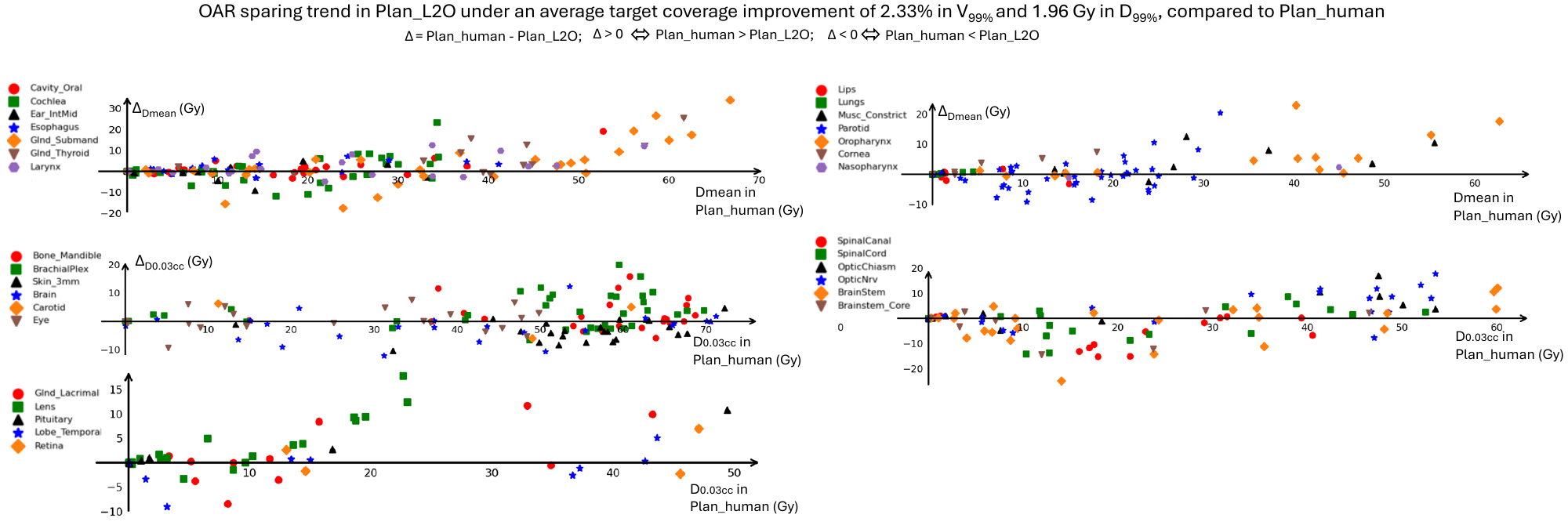}
\caption{Trend of OAR sparing improvements observed in Plan\_L2O under an average target coverage improvement of 2.33\% in $V_{99\%}$ and 1.96~Gy in $D_{99\%}$, relative to Plan\_human across 25 test patients.}
\label{fig5}
\end{figure}

Figure~\ref{fig5} visualizes the trend of OAR sparing improvements achieved by Plan\_L2O, with target coverage increasing by an average of 2.33\% in $V_{99\%}$ and 1.96~Gy in $D_{99\%}$ across all 25 test patients.
As illustrated, does received by OARs are more likely to be reduced in high-dose regions, suggesting that the proposed method preferentially spares OARs exposed to higher radiation without compromising target coverage.

These results show that plans generated by the proposed learning-driven automatic treatment planning framework achieves comparable or better OAR sparing with superior target coverage when compared with human-generated plans. 
For each patient, five plans can be produced in an average planning time of 2.55 hours, demonstrating that the proposed approach is both effective and efficient, capable of producing high-quality plans within clinically acceptable planning times.

\section{DISCUSSION}

Automatic treatment planning is an iterative process of objective parameter adjustment and inverse optimization. 
The overall performance of this process depends on the intelligence of the adjustment mechanism and the efficiency and effectiveness of the inverse optimizer.
While DRL has been extensively explored for the automatic adjustment of objective parameters, inverse optimization remains dependent on traditional time-consuming gradient-based methods. 
Accelerating optimization and improving its accuracy would enable more effective and efficient exploration of the objective parameter space, thereby enhancing the planning model’s capacity to identify optimal solutions.
Here, the high-quality plans generated in reasonable planning times demonstrate the superior effectiveness and efficiency of our learning-driven solution for proton PBS treatment planning.
The superior performance can potentially be attributed to the fact that our L2O optimizer performs computations on GPUs, and by learning from task-specific data distributions, it can identify more effective update steps than traditional theory-driven gradient-based optimizers.


This pilot study has the following limitations and areas for improvement. 
First, physicians may have different treatment preferences from general consensus in radiation oncology on which our policy network is trained.
As seen in Case B08, where the physician directed human planner to spare Glnd\_Submand\_R while the virtual planner did not receive such physician guidance, the overlapping volume between the target and Glnd\_Submand\_R was categorized differently and in turn received different doses.
Therefore, a human-in-the-loop mechanism that allows the radiation oncologist to interact and adjust the optimization parameters based on physician preferences for the particular patient is desirable. 
Such physician-directed treatment planning could be realized by using LLMs such as ChatGPT to customize overlapping volume definition or specify specific dose requirements. 
For future work, we will explore integrating our planning framework with LLM-based agents equipped with tool-use capabilities. 

In addition, the current implementation is constrained by GPU memory, limiting simultaneous optimization to 25,000 spots and restricting the L2O network to about 10 million parameters. 
Since this limitation is hardware-dependent rather than algorithmic, the framework is inherently scalable and can be readily extended by provisioning additional GPU memory. 
Modern billion-parameter LLMs already support million-token context windows, underscoring the strong potential to expand L2O methods to much larger optimization problems. 
While the present lightweight model is sufficient to generate complex treatment plans with human-level quality for most clinical cases, increasing spot capacity and enabling larger networks could further enhance plan quality, ultimately paving the way for broader clinical application.

\section{CONCLUSION}
This study introduces a novel and promising direction for inverse optimization in proton PBS treatment planning by incorporating, for the first time, long-context processing techniques into a Transformer-based L2O framework.
By learning update strategies from task-specific data distributions, the proposed L2O optimizer improves both effectiveness and efficiency, achieving gains of 22.97\% and 36.41\%, respectively, over conventional rule-based optimization techniques. 
It is evidenced that our L2O optimizer effectively resolves the scalability limitations of existing L2O methods, generalizing L2O to large-scale inverse optimization problems. 
Furthermore, when integrated into a PPO-based virtual planner along with a Swin~UnetR dose predictor, the overall framework is capable of generating treatment plans with improved or comparable OAR sparing and superior target coverage compared to human-generated clinical plans, demonstrating its capability of consistently producing high-quality plans within clinically acceptable planning times, i.e., an average of 2.55 hours, across diverse treatment scenarios. This highlights the potential of proposed framework for practical clinical use.

\newpage
\appendix
\section{Dose limits for initializing the optimization and OAR-specific allowable dose intervals used in PPO}
\label{app0:table}

\begin{table}[htbp]
\centering
\scriptsize
\caption{ Dose limits used to initialize the optimization together with predictions from Swin~UnetR, and OAR-specific allowable dose intervals used by the reward function in PPO.}
\begin{tabular}{|c|c|c|c|c|c|}
\hline
OAR             & \makecell[c]{$D_{clinic}$ \\(Dmax objectives)} & dose intervals & OAR             & \makecell[c]{$D_{clinic}$ \\(Dmean objectives)} & dose intervals \\ \hline
SpinalCord      & 45           &[0, 10]            & Ear\_IntMid     & 40            & [0, 15]            \\ \hline
OpticChiasm     & 54           & [0, 20]            & Cavity\_Oral    & 35            & [0, 10]            \\ \hline
OpticNerve      & 54           & [0, 20]            & Glnd\_Lacrimal  & 35            & [0, 15]            \\ \hline
BrainStem       & 54           & [0, 20]            & Oropharynx      & 45            & [0, 20]            \\ \hline
BrainStem\_Core & 54           & [0, 20]            & Glnd\_Submand   & 45            & [0, 20]            \\ \hline
Bone\_Mandible  & 60           & [0, 20]            & Glnd\_Thyroid   & 45            & [0, 20]            \\ \hline
BrachialPlex    & 63           & [0, 20]            & Larynx          & 45            & [0, 20]            \\ \hline
Brain           & 60           & [0, 20]            & Lips            & 26            & [0, 5]             \\ \hline
Lens            & 10           & [0, 0]             & Cornea          & 45            & [0, 20]            \\ \hline
Cochlea         & 45           & [0, 10]            & Lungs           & 20            & [0, 5]             \\ \hline
Pituitary       & 56           & [0, 20]            & Musc\_Constrict & 45            & [0, 20]            \\ \hline
Esophagus       & 60           & [0, 20]            & Nasopharynx     & 45            & [0, 20]            \\ \hline
Lobe\_Temporal  & 60           & [0, 20]            & Parotid         & 30          & [0, 10]            \\ \hline
Retina          & 50           & [0, 15]            &                 &               &               \\ \hline
VocalCords      & 60           & [0, 20]            &                 &               &               \\ \hline
Carotid         & 54           & [0, 20]            &                 &               &               \\ \hline
SpinalCanal     & 45           & [0, 10]            &                 &               &               \\ \hline
Skin            & 54           & [0, 20]            &                 &               &               \\ \hline
Eye             & 50           & [0, 15]            &                 &               &               \\ \hline
\end{tabular}
\label{tab:tab1}
\end{table}

\section{Normalization of the L2O network inputs}
\label{app1:norm}

Given $N$ proton spots in the plan and $K$ weighted components in the objective function in Equation~\ref{eq:eq1}, the input is a matrix with shape $[N, K+2]$. At timestep $t$, we normalize the gradient $g_t$, $g_{kt}$ and the momentum $m_t$ with:
\begin{align}
\tilde{g_t} = &\frac{g_t}{\sqrt{\hat{v_t}} + \epsilon} ;\quad   \tilde{g_{kt}} = \frac{g_{kt}}{\sqrt{\hat{v_t}} + \epsilon};\quad  \tilde{m_t} = \frac{\hat{m_t}}{\sqrt{\hat{v_t}} + \epsilon} \notag \\
\hat{m_t} &=\frac{m_t}{1-\beta_1^t} ; \quad m_t = \beta_1 \, m_{t-1}  + (1-\beta_1) \, g_t  \label{eq:eq3}\\
\hat{v_t} &= \frac{v_t}{1-\beta_2^t};   \quad v_t = \beta_2 \, v_{t-1} + (1-\beta_2) \, g_t^2 \notag
\end{align}
where $g_{kt}$ represents the gradient computed from the $k$-th component of the objective function, $\epsilon$ is a small constant $1\times10^{-8}$, and $\beta_1$ and ${\beta_2}$ are set to 0.9 and 0.99, respectively. $v_0$ and $m_0$ are gradient and momentum at timestep 0, and set to all-0 matrices. 

\section{Loss function used to train proposed L2O network}
\label{app2:loss}

The 6-layer Transformer-based L2O network is configured with a hidden size of 256, an intermediate size of 512, and eight attention heads. In the GQA mechanism, the number of key and value heads is set to four.
Moreover, to enhance the training stability, assuming the objective losses are aggregated to update the L2O network every $T$ steps, the weighting coefficient $w_t$ in Equation~\ref{eq:eq2} is designed as:
\begin{equation}
w_t = min\biggl(1.0, \frac{n_{step}}{20(\frac{n_{epoch}}{10})^2}\biggr) 
\label{eq:eq4}
\end{equation}
at the $n_{step}$ step of the $n_{epoch}$ epoch. Subsequently, the loss used to train the network at the $n_{step}$ step can be rewritten as:
\begin{equation}
L(\phi) = \mathbb{E}_{f\in \Gamma}\Biggl[\sum_{t=1}^T min\biggl(1.0, \frac{n_{step}}{20(\frac{n_{epoch}}{10})^2}\biggr) f(X_t)\Biggr], 
\label{eq:eq5}
\end{equation}
where $\Gamma$ is a small batch of patients, and $f(\cdot)$ is the objective function defined in Equation~\ref{eq:eq1}.

\vspace{1cm}
\noindent
{\bf Declaration of generative AI and AI-assisted technologies in the manuscript preparation process}
During the preparation of this work the authors used OpenAI ChatGPT for language refinement. 
After using this tool/service, the authors reviewed and edited the content as needed and take full responsibility for the content of the published article.




\end{document}